\documentclass[10pt,twocolumn,letterpaper]{article}

\usepackage{iccv}              
\usepackage{algorithm}
\usepackage{amssymb,amsfonts}
\usepackage{amsthm}
\usepackage{algorithmicx}
\usepackage{multirow}
\usepackage{wrapfig}

\definecolor{iccvblue}{rgb}{0.21,0.49,0.74}
\usepackage[pagebackref,breaklinks,colorlinks,allcolors=iccvblue]{hyperref}

\def\paperID{5890} 
\def\confName{ICCV}
\def\confYear{2025}

\title{How Do Optical Flow and Textual Prompts Collaborate to Assist in Audio-Visual Semantic Segmentation?}

\author{$\text{Yujian Lee}^{1,2,\dagger}$, \and $\text{Peng Gao}^{1,2,\dagger}$, \and $\text{Yongqi Xu}^{3}$, \and $\text{Wentao Fan}^{1,2,*}$ \and
$^1\text{Hong Kong Baptist University, Hong Kong, China}$\\
$^2\text{Guangdong Provincial/Zhuhai Key Laboratory IRADS}$\\ and Department of Computer Science, Beijing Normal-Hong Kong Baptist University, Zhuhai, China\\
$^3\text{Peking University, Shenzhen Graduate School, Shenzhen, China}$\\
{\tt\small \{yujianlee1119, gaopeng1225\}@gmail.com, xuyongqi@stu.pku.edu.cn, wentaofan@uic.edu.cn}
\thanks{\textbf{$^\dagger\text{Equal Contribution}$,$^*\text{Corresponding author}$.} The completion of this work was supported by the National Natural Science Foundation of China (62276106), Guangdong Basic and Applied Basic Research Foundation (2024A1515011767), the Guangdong Provincial Key Laboratory IRADS (2022B1212010006), and the Guangdong Higher Education Upgrading Plan (2021-2025) with No. of  (2024GXJK695, 2024KTSCX222).}
}

\begin{document}
\maketitle
\begin{abstract}
Audio-visual semantic segmentation (AVSS) represents an extension of the audio-visual segmentation (AVS) task, necessitating a semantic understanding of audio-visual scenes beyond merely identifying sound-emitting objects at the visual pixel level. Contrary to a previous methodology, by decomposing the AVSS task into two discrete subtasks by initially providing a prompted segmentation mask to facilitate subsequent semantic analysis, our approach innovates on this foundational strategy. We introduce a novel collaborative framework, \textit{S}tepping \textit{S}tone \textit{P}lus (SSP), which integrates optical flow and textual prompts to assist the segmentation process. In scenarios where sound sources frequently coexist with moving objects, our pre-mask technique leverages optical flow to capture motion dynamics, providing essential temporal context for precise segmentation. To address the challenge posed by stationary sound-emitting objects, such as alarm clocks, SSP incorporates two specific textual prompts: one identifies the category of the sound-emitting object, and the other provides a broader description of the scene. Additionally, we implement a visual-textual alignment module (VTA) to facilitate cross-modal integration, delivering more coherent and contextually relevant semantic interpretations. Our training regimen involves a post-mask technique aimed at compelling the model to learn the diagram of the optical flow. Experimental results demonstrate that SSP outperforms existing AVS methods, delivering efficient and precise segmentation results.
\end{abstract}    
\section{Introduction}
\label{sec:intro}
The nature of the audio-visual segmentation (AVS) task is to pinpoint sound-emitting objects in the visual domain using auditory cues \cite{arandjelovic2018objects}. Prior to the emergence of AVS, audio-visual localization (AVL) provides general sound source locations \cite{tian2018audio, wu2019dual}. However, AVL lacks the precision needed to identify specific sound-producing objects. AVS addresses this limitation by offering pixel-level labels for such acoustic entity \cite{zhou2022audio}. Nevertheless, identifying these objects from the visual scene without prior knowledge of their nature proves impractical in daily life. Therefore, audio-visual semantic segmentation (AVSS) enhances this process by predicting semantic labels for the objects \cite{zhou2024audio}. By integrating semantic understanding, AVSS improves machine perception of the auditory environment.
\begin{figure*}[!t]
  \centering
  \makebox[1\textwidth]{\includegraphics[scale=0.56]{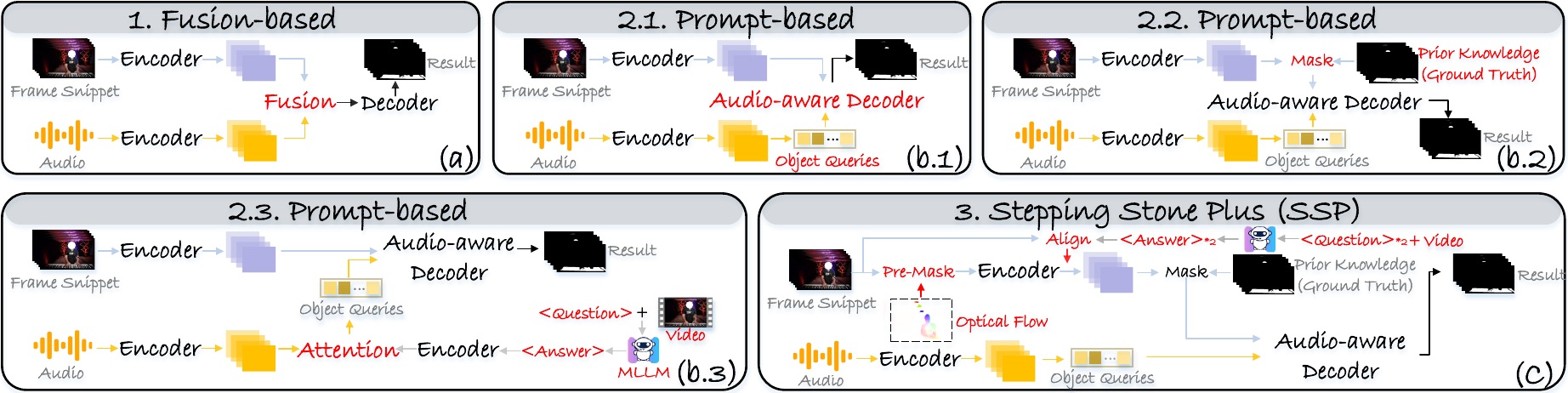}}
  \caption{Network architecture comparison, red text is utilized to denote the innovative modules within the respective methods. Fusion-based methods in (a), three categories of prompt-based methods (object queries, mask, and textual prompt in (b.1), (b.2), and (b.3), respectively), and our proposed method in (c). SSP introduces optical flow as an extra prompt alongside the pre-masking technique to the original image to achieve better segmentation mask results, and the incorporation of two textual prompts compensating for optical flow. The alignment module delivers more robust feature maps for visual inputs.}\label{intro}
\end{figure*}

The design of the existing AVS approaches can be broadly categorized into fusion-based methods \cite{zhou2022audio, huang2023discovering, li2023catr, liu2023audio, gao2024avsegformer, li2025audio, pian2025continual, park5132584save} and prompt-based methods \cite{mo2023av, yan2024referred, wang2024prompting, wang2025can, ma2024stepping, gong2025avs, wang2025transformer, yang2024cooperation}. (1). Fusion-based methods in Fig. \ref{intro}(a) combine auditory and visual modal information into a unified representation to accomplish the AVS task. These methods typically utilize network architectures such as convolutional neural networks (CNNs) and transformers to facilitate effective modality fusion. A prominent example is AVSBench \cite{zhou2022audio}, where visual information is first encoded by an encoder to generate multi-scale visual representations. Subsequently, audio features are mapped to these multi-scale representations, enabling the effective fusion of cross-modal information \cite{qingyun2021cross}. (2). Prompt-based methods in Fig. \ref{intro}(b) can be further subdivided into three categories (Fig. \ref{intro}(b.1), Fig. \ref{intro}(b.2), and Fig. \ref{intro}(b.3)): (2.1). One category involves object queries feature embeddings generated from auditory features \cite{gong2025avs, wang2024prompting, wang2025can, wang2025transformer, yang2024cooperation} or (2.2). Masks created from visual features \cite{ma2024stepping, mo2023av}, both derived from the original data. (2.3). The other category encompasses additional prior contextual knowledge, textual prompts \cite{liu2024annotation, wang2025can, yan2024referred, yang2024cooperation}, generated by multimodal large language models (MLLMs) \cite{huang2024large, wu2023multimodal}, which can be a description to the image snippets or a word indicating the class of the potential sounding object.  

The limitations of the existing approaches underscore a gap in the field. Specifically, in (1), critical information from either input will be lost if the model is not optimally designed to balance contributions from both modalities \cite{gong2025avs}. In (2), particularly concerning the textual prompts, which are static representations, offering a macro-level description of the video rather than a dynamic depiction that aligns with the audio's sounds. Moreover, the prompts generated in (2.1) and (2.2) are produced internally by the model, enabling them to actively participate in subsequent interactions across both audio and visual modalities. Conversely, in method (2.3), the MLLM generates textual outputs, the information will then be encoded by the contrastive language-image pre-training model (CLIP) \cite{radford2021learning}. This process leads to the encoding of distinct modality signals using separate encoders, resulting in isolated latent spaces, and discrepancies between modalities will arise \cite{baltruvsaitis2018multimodal, pan2010cross, yu2025clipceil}.

\begin{figure}[!t]
  \centering
  \makebox[0.5\textwidth]{\includegraphics[scale=0.6]{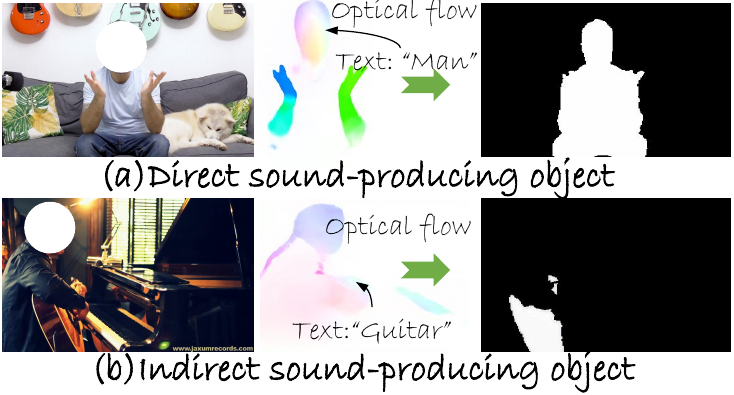}}
  \caption{Optical flow carries information about sound-emitting objects, which can be either (a) the dynamic object itself (the man) or (b) indirectly related to stationary object (the guitar).}\label{visrb}
  \vspace{-0.3cm}
\end{figure}


Utilizing the previous \cite{ma2024stepping, 11209970} as our baseline and addressing the limitations of static textual prompts that lack spatial information, we propose a novel framework in Fig. \ref{intro}(c), Stepping Stone Plus (SSP). To the best of our knowledge, we are the first to have the optical flow as an auxiliary mask prompt for the task, shown in Fig. \ref{visrb}. Such a prompt can enhance the generation of segmentation masks by having the proposed pre-mask technique, as sound sources are often associated with the movement of objects. Optical flow highlights moving objects in Fig. \ref{visrb}(a) and provides valuable prior knowledge regarding visual information \cite{hu2025mona, yuan2024unsamflow, jaegle2021perceiver, liu2024audio}. However, there are instances when sound is produced by stationary objects in Fig. \ref{visrb}(b) \cite{qiu2024time}. To address this, we employ an off-the-shelf MLLM (MiniCPM-o-2.6 \cite{yao2024minicpm}) to introduce two types of textual prompts into the complete video input. One prompt identifies the type of sound-emitting object, while the other describes the entire scene. These supplementary text-assisted inputs effectively address the aforementioned scenarios where objects remain stationary. Moreover, recognizing the necessity of encoding textual information through the encoder, we develop a visual-textual alignment module (VTA) to have seamless integration of visual and textual data. This design employs the bidirectional encoder representations from transformers (BERT) \cite{kenton2019bert} as the backbone, overcoming the challenges posed by disparate encoders. In our training objectives, we incorporate a post-mask technique based on a mask prediction loss, to further reinforce the model's acquisition from the result of the pre-mask technique. By adopting this holistic framework, SSP significantly boosts the efficacy of segmentation. The main contributions of SSP are four-fold:
\begin{itemize}
  \item \textbf{Pre-mask technique with optical flow:} SSP uniquely integrates optical flow as a prompt, combining it with the ground truth mask within the pre-masking technique. The return output applied to the original image can better refine the generation of segmentation masks.
  \item \textbf{Textual prompts:} Two distinct textual prompts are utilized, with one addressing scenarios where the sound originates from stationary objects and the other enriches the overall visual semantic understanding.
  \item \textbf{Visual-textual alignment module (VTA):} VTA integrates the visual and textual information effectively, returning unified cross modalities representations.
  \item \textbf{Training objective:} The output of the pre-mask technique provides prior knowledge; thus, we apply an extra constraint, a post-mask technique, to have the model assimilate this knowledge.
\end{itemize}
\section{Related Work}
\label{sec:related}
\subsection{Assisting AVS through prompts}
In the context of AVS tasks, prompts (i.e. object queries from auditory features, textual prompts from MLLMs, and visual mask features) can assist in bridging the gap between audio and visual information to boost the segmentation efficacy. (1). The utilization of object queries in \cite{gong2025avs, wang2024prompting, wang2025can, wang2025transformer} involves mapping these queries from original data to specific categories, which then interact with visual features to yield segmented objects and corresponding mask categories. This process enriches semantic representation, enabling dynamic interactions with objects over time \cite{huang2023discovering}. (2). With the emergence of MLLMs, the introduction of textual guidance into the AVS task shows great promise \cite{zhang2401mm, achiam2023gpt, lund2023chatting}. Approaches such as those in \cite{liu2024annotation, wang2025can, yan2024referred} leverage MLLMs to identify potential sound-producing objects, integrating these insights with audio features to generate object queries similar to (1). (3). Mask generated from the integration of the encoded visual features with the ground truth mask labels (excluded during the inference stage) in \cite{ma2024stepping} serve as crucial stepping stones, that can provide a shortcut to help delineate object boundaries and enhance segmentation accuracy \cite{mostafa2024yolo}. While these approaches have demonstrated performance improvements, the enhancements are often modest. To address this, we introduce dynamic prompts through optical flow, which allows the model to focus more acutely on relevant areas within the visual domain, facilitating a more streamlined and effective segmentation workflow. Additionally, the complementary textual prompt can mitigate the limitations of optical flow, further enriching the model's understanding in the AVS task.

\subsection{Cross modalities alignment}
When a model integrates supplementary information, the lack of appropriate organization can lead to redundancy \cite{ye2024x, lin2022adapt, chen2023rethinking}, even if the introduced information possesses intrinsic value. This underscores the necessity for effective modality alignment. Existing AVS approaches incorporate textual prompts that offer substantial prior knowledge and can assist the segmentation process \cite{liu2024annotation, wang2025can, yan2024referred, yang2024cooperation}. However, encoding the textual content and subsequently fusing, concatenating, or employing transformers \cite{han2021transformer} with the original feature embeddings for later segmentation poses a risk of losing critical data features, thereby diminishing the efficacy of the introduced textual prompts \cite{ma2024learning, wang2024improving}. To adeptly tackle the challenges associated with modality alignment, we propose a cross-modal alignment module utilizing BERT as the backbone \cite{li2023blip}. By inputting cross-modal features in their original, unencoded form, the model generates a unified feature embedding. This methodology not only preserves the integrity of the original data but also facilitates seamless integration, thereby substantially improving the alignment between modalities and harnessing the full potential of the available information \cite{li2022blip}.



\section{Preliminaries}
\label{sec:preliminary}
This section presents pre-definitions of feature extractor backbones and the baseline method AAVS \cite{ma2024stepping}. 

\textbf{Preprocess.} Before visual data is encoded to the model, it will be preprocessed by Masked-attention Mask Transformer (Mask2Former) \cite{cheng2022masked}, returns the normalized and resized image $\text{V}^\mathcal{T}$ data, without any other extra operation.

\textbf{Encoders.} Given a video clip comprising $\mathcal{T}$ frames, we employ the pre-trained visual encoder as described in \cite{zhou2022audio, li2023catr} to extract visual context $\text{V}^\mathcal{T}$ across four distinct scales: 1/4, 1/8, 1/16, and 1/32 of the original image dimensions. This results in visual feature representations denoted as $\mathcal{Z}_\text{V}^{\mathcal{T}\times \mathcal{C}_\text{v}\times \mathcal{H}_\text{v} \times \mathcal{W}_\text{v}}$, where $\text{v} \in \{1,2,3,4\}$ and $\mathcal{C}_\text{v} \in \{128,256,512,1024\}$. For the audio input, we utilize the pre-trained VGGish model \cite{simonyan2014very} to process the spectrogram, thereby extracting audio features represented as $\mathcal{Z}_\text{A}^{\mathcal{T} \times \mathcal{C}_\text{A}}$ ($\mathcal{C}_\text{A}$ = 256). Object queries $\mathcal{Q}_\text{Obj}^{\mathcal{N\times C_\text{A}}}$ is the learnable parameter, designed to map audio features into  $\mathcal{N}$ distinct object categories. The cosine similarity and Einstein summation convention applied between $\mathcal{Z}_\text{A}$ and $\mathcal{Q}_\text{Obj}$ to finish this mapping, the result $\mathcal{Q}_\text{Obj}'^{\mathcal{N\times T\times C_\text{A}}}$ will be interacting with the visual features for the decoding process.

\textbf{Decoder.} The feature pyramid network (FPN) \cite{lin2017feature} is a visual pixel decoder to aggregate $\mathcal{Z}_\text{V}$ in various resolutions to $\mathcal{Z}_\text{V}'^{\mathcal{T}\times \mathcal{C}_\text{v}'\times \mathcal{H}_\text{v}' \times \mathcal{W}_\text{v}'}$, with $\mathcal{C}_\text{v}'$ unified to be 256. The audio-aware decoder based on the Mask2Former, takes $\mathcal{Z}_\text{V}'$ and $\mathcal{Q}_\text{Obj}'$ as inputs, by combining multi-scale visual features and object queries to realize weighting and generation of segmentation masks through the attention mechanism.

\textbf{Baseline.} The baseline AAVS \cite{ma2024stepping} presented in Fig. \ref{intro}(b.3) conceptualizes the AVSS task as an integration of AVS and semantic segmentation (SS). In the first stage, the AVS model is trained using binary mask labels that emphasize sound source localization. In the subsequent stage, the SS supervised by the semantic mask labels leverages the mask results obtained from the AVS, enhanced by providing prior knowledge through the ground truth (GT) mask label $\mathcal{M}_\text{GT}$. This is integrated with the final hidden state of the decoded visual features $\mathcal{Z}_\text{V}'^{\mathcal{T}\times \mathcal{C}_\text{4}'\times \mathcal{H}_\text{4}' \times \mathcal{W}_\text{4}'}$ through interpolation, illustrated in Fig. \ref{method}. Notably, $\mathcal{M}_\text{GT}$ is employed during training but excluded during testing.

\section{Methodology}
\label{sec:method}
\begin{figure*}[!t]
  \centering
  \makebox[1\textwidth]{\includegraphics[scale=0.7]{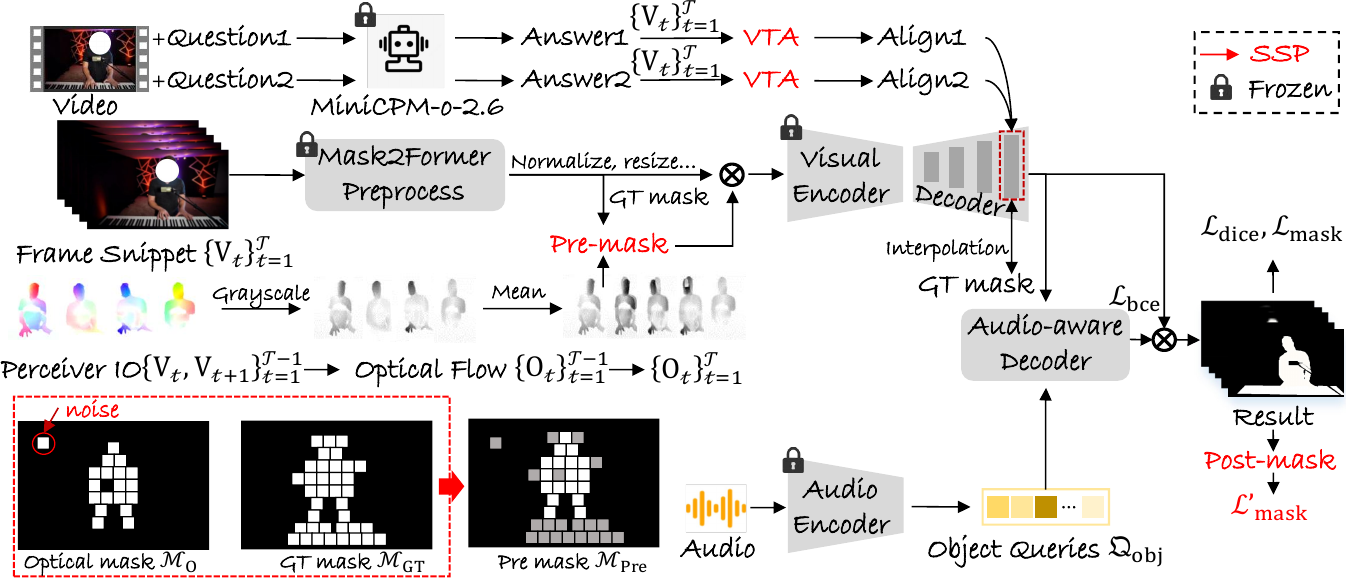}}
  \caption{The workflow of the SSP begins by incorporating optical flow $\text{O}^\mathcal{T}$ as a prompt. We then generate the optical flow mask, denoted as $\mathcal{M}_\text{O}$, to interact with the GT mask $\mathcal{M}_\text{GT}$ within the pre-mask technique. This process returns the output $\mathcal{M}_\text{Pre}$, which is then multiplied with the visual features to identify moving and sound-producing objects; however, scenarios exist the sound-producing but stationary objects (the gray region in $\mathcal{M}_\text{Pre}$). To account for this, we introduce two textual prompts: $\text{A}_1$, which offers a comprehensive understanding of the scene and serves as the foundation for $\text{A}_2$, and $\text{A}_2$, indicates potential stationary objects capable of producing sound. The VTA facilitates the alignment process across modalities. The two aligned features, $\text{Align}_1$ and $\text{Align}_2$, are integrated with feature embeddings from the last hidden state of the visual decoder. In our training objective, we employ an additional mask prediction loss, $\mathcal{L}'_\text{mask}$, alongside the post-mask technique illustrated in Fig. \ref{loss}. This approach establishes the ground truth and compels the model to learn features that are both dynamic and sound-related.} \label{method}
\end{figure*}
We delineate the modules of our proposed SSP framework in Fig. \ref{method}. We begin with the pre-mask technique, the generation of textual prompts, followed by the VTA module, and outline our training objective with a post-mask technique. By deconstructing the components of SSP, we elucidate how the interplay among the prompts facilitates dual-prompt assistance, thereby enhancing model performance.

\subsection{Pre-mask with Optical Flow}
The pre-mask technique is the core SSP, with its essence lies in preserving potential sound-producing objects through the integration of the optical mask $\mathcal{M}_\text{O}$ and the GT mask $\mathcal{M}_\text{GT}$ to assist the efficacy of subsequent segmentation. $\mathcal{M}_\text{GT}$ is utilized during training and excluded in testing. 

Before obtaining the result of the pre-mask technique $\mathcal{M}_\text{Pre}$, we clarify the acquisition of $\mathcal{M}_\text{O}$. We utilize the Perceiver IO \cite{jaegle2021perceiver} to obtain the optical flow data, denoted as $\text{O}^{\mathcal{T}-1}$. The optical flow estimates the motion of the dynamic objects between the last frame and the current frame; however, the result does not directly yield the mask for the current frame. We convert $\text{O}^{\mathcal{T}-1}$ into grayscale images, then have the following transformation:

\begin{align}
    \text{O}^{\mathcal{T}}=\text{Stack}
    \begin{cases} 
        \text{O}^{1}; & \\ 
        \text{Mean}(\text{O}^{t},\text{O}^{t+1}), & t = 1,2,...,\mathcal{T}-1; \\ 
        \text{O}^{\mathcal{T}}. & 
    \end{cases}
\end{align}
Computing the mean value of the adjacent frame can smooth out optical bias and enhance the visual consistency with the original images \cite{dosovitskiy2015flownet, ilg2017flownet}. We then convert $\text{O}^{\mathcal{T}}$ into the binary optical mask $\mathcal{M}_\text{O}$, where the blank areas are assigned a value of 0, while regions containing information are designated with a value of 1. 

We aim to ensure that the mask $\mathcal{M}_\text{O}$ contains as much accurate information as possible. However, since the data derived from the optical flow may not always be reliable, we first compute the intersection of $\mathcal{M}_\text{O}$ with the ground truth mask $\mathcal{M}_\text{GT}$. This step guarantees that the resulting mask includes a partially segmented output while filtering out background information. For the remaining pixels present in either $\mathcal{M}_\text{O}$ or $\mathcal{M}_\text{GT}$, we categorize them as uncertain, as depicted by the gray region in $\mathcal{M}_\text{Pre}$ in Fig. \ref{method}, they either be the noise pixel or the pixel that can not be detected in $\mathcal{M}_\text{O}$. The result of $\mathcal{M}_\text{Pre}$ can be defined as:

\begin{align}
    \mathcal{M}_\text{Pre}(h,w)&=
    \begin{cases} 
         1,\quad\quad (\mathcal{M}_\text{O} \cap \mathcal{M}_\text{GT})(h,w) = 1; \\ 
         0.5,\quad(\mathcal{M}_\text{O} \cup \mathcal{M}_\text{GT}- \mathcal{M}_\text{O}\\ \quad\quad\quad \cap \mathcal{M}_\text{GT})(h,w) = 1; \\ 
        0,\quad\quad (\mathcal{M}_\text{O} \cap \mathcal{M}_\text{GT})(h,w) = 0;\\
    \end{cases}\nonumber \\
    &h,w \in \mathcal{H}, \mathcal{W},
\end{align}
where 1 signifies strong agreement of segmented objects. A value of 0.5 suggests the presence of uncertainty. The value of 0 represents the background, containing blank information. $\mathcal{M}_\text{Pre}$ will then be multiplied to $\text{V}^\mathcal{T}$.

\subsection{Textual Prompts}
To compensate for the scenario of stationary objects will have the potential to produce sound (i.e. the gray region of $\mathcal{M}_\text{Pre}$ in Fig. \ref{method}, where the man is singing with the piano playing, sound-producing objects are the man and the piano), we incorporate two distinct types of textual prompts generated by the MiniCPM-o-2.6 \cite{yao2024minicpm} to assist the task. Input a video and a specific question, it returns a textual caption. In Fig. \ref{method}, we have designed two questions, denoted as $\text{Q}_1$, $\text{Q}_2$, with their respective outputs $\text{A}_1$ and $\text{A}_2$. Notably, $\text{A}_1$ serves as the foundation for 
$\text{A}_2$, establishing a hierarchical relationship between the two responses that enriches the understanding of the scene.

$\text{A}_1$ is the description of the video, thus, $\text{Q}_1$ is set to be \textless\textit{Describe the video, give every object's location with their characteristic, with a format of ``A NOUN VERB NOUN", for example: ``a man plays a guitar".}\textgreater. The content of $\text{A}_1$ is \textless\textit{A man is playing a piano, wearing a dark gray t-shirt, produces musical notes, and the microphone near the person's mouth suggests singing or speaking.}\textgreater. $\text{Q}_2$ is to search the potential sound-producing objects, with $\text{A}_1$ provides the necessary semantic context, therefore, by defining $\text{Q}_2$ to be \textless\textit{According to the video, with $\text{A}_1$, give the NOUN word that potentially produce sound.}\textgreater. The answer $\text{A}_2$ \textless\textit{piano, microphone, man.}\textgreater can easily spot out the potential sound-producing objects.

\subsection{Visual-Textual Alignment}
Only when the two textual prompts are utilized effectively can it fulfill the gray region and filter out noise in $\mathcal{M}_\text{Pre}$, thus, we introduce the VTA module in Algorithm \ref{VTA}, applying BERT as the backbone.

\begin{algorithm}[!t]
\caption{Visual-Textual Alignment.}
\label{VTA}
\begin{algorithmic}[1] 
\State \textbf{Input:} $\text{A}_1$, $\text{A}_2$, $\text{V}^\mathcal{T}$, attention mask for visual inputs ($\text{Attmask}^\text{V}$).
\State \textbf{Output:} $\text{Align}_1$, $\text{Allign}_2$.
\State Parallel operation for $\text{A}_1$, $\text{A}_2$, using A and Align for the process instead.
\State $\mathcal{Z}_\text{VTA}^\text{V}$ = MLP(CLIP($\text{V}^\mathcal{T}$)),\\ $\mathcal{Z}_\text{VTA}^\text{A}$, $\mathcal{Z}_\text{VTA}^\text{A-Att}$  = MLP(Tokenizer(A)).
\State Attmask-Unify = $\text{Cat}(\mathcal{Z}_\text{VTA}^\text{A-Att},\text{Attmask}^\text{V})$.
\State Align = BERT$\left(\mathcal{Z}_\text{VTA}^\text{A}, \text{Attmask-Unify}, \mathcal{Z}_\text{VTA}^\text{V}, \text{Attmask}^\text{V} \right)$,

\State Align = BERT$\left(\mathcal{Z}_\text{VTA}^\text{A}, \text{Attmask-Unify}, \text{Align} \right)$.

\State \textbf{return} Normalize(MLP(Align))

\end{algorithmic}
\end{algorithm}

In Algorithm \ref{VTA}, visual data is encoded using CLIP, while text data is processed by BLIP \cite{li2023blip} to generate tokens. Prior to utilizing BERT on the aligned features, merging the attention masks of the two modalities into a single mask facilitates uniform treatment of visual and textual data, thereby enhancing overall adaptability. BERT is subsequently invoked a second time to further refine the text features, capitalizing on the fusion features generated in the initial pass. This step is pivotal in augmenting the text representation with prior context, the synthesized visual and textual information. This integration empowers VTA to discern the intricate relationships between the textual content and its corresponding image information more effectively. Ultimately, VTA returns the aligned features, represented as $\text{Align}_1$ and $\text{Align}_2$, encapsulating the enhanced understanding derived from the multimodal interplay. The last step of VTA is 

\begin{align}
    \mathcal{Z}'^{\mathcal{T}\times \mathcal{C}'_{\text{4}} \times \mathcal{H}'_{\text{4}} \times \mathcal{W}'_{\text{4}}}_\text{V} = \text{Normalize}(&\mathcal{Z}'^{\mathcal{T}\times \mathcal{C}'_{\text{4}} \times \mathcal{H}'_{\text{4}} \times \mathcal{W}'_{\text{4}}}_\text{V} + \text{Align}_1 \nonumber \\
    & + \text{Align}_2),
\end{align}
integrating with two aligned features compensating the gray region as well as filtering out the noise in $\mathcal{M}_\text{Pre}$ in Fig. \ref{method}.

\subsection{Training Objective}
\begin{figure}[!t]
  \centering
  \makebox[0.5\textwidth]{\includegraphics[scale=0.75]{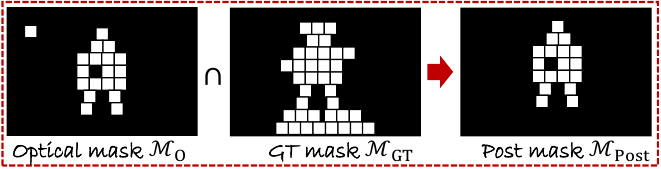}}
  \caption{Ground truth label preparation for the post-mask technique, by having the intersection between the optical flow mask $\mathcal{M}_\text{O}$ and the original ground truth mask $\mathcal{M}_\text{GT}$.}\label{loss}
\end{figure}
In the training process, we adopt binary cross-entropy loss $\mathcal{L}_\text{mask}$ and the dice loss $\mathcal{L}_\text{dice}$ \cite{cheng2022masked} to enhance the accuracy of mask predictions. Besides, the mask classification loss $\mathcal{L}_\text{bce}$ is applied to achieve the objective of classification of the masked objects. The composite loss function for AVS is 

\begin{align}
\mathcal{L}_\text{AVS} = \lambda_\text{mask}\mathcal{L}_\text{mask} + \lambda_\text{dice}\mathcal{L}_\text{dice} + \lambda_\text{bce}*\mathcal{L}_\text{bce},
\end{align}
where the parameters $\lambda_\text{mask}$, $\lambda_\text{dice}$, and $\lambda_\text{bce}$ are set to 5, 5, and 2, respectively, serving as weights to balance the influence of each loss component.

Rather than allowing the model to overly depend on the $\mathcal{M}_\text{GT}$, which will be excluded during testing, we introduce an additional mask prediction loss, denoted as $\mathcal{L}'_\text{mask}$. The label for $\mathcal{L}'_\text{mask}$ is defined as $\mathcal{M}_\text{Post}$, which is the intersection between $\mathcal{M}_\text{O}$ and $\mathcal{M}_\text{GT}$, in Fig. \ref{loss}. This additional loss is designed to encourage the model to learn features that are dynamic and sound-related, improving its generalization ability without relying on ground truth during inference. The granularity of $\mathcal{L}'_\text{mask}$ is coarser than that of $\mathcal{L}_\text{mask}$, which allows for a higher fault tolerance rate. If the model's predictions fall outside this intersection, it indicates a significant error. To intensify the penalty for such discrepancies, the hyperparameter $\lambda'_\text{mask}$ is set to 10. The overall loss function of SSP is:

\begin{align}
    \mathcal{L} = \mathcal{L}_\text{AVS} + \lambda'_\text{mask}*\mathcal{L}'_\text{mask}.
\end{align}

\section{Experiment}
\label{sec:experiment}
\subsection{Experimental Settings}
\textbf{Experiments.} Four types of experiments, including (1) comparative study, (2) ablation study, (3) qualitative analysis, and (4) visualization results are designed to comprehensively evaluate the efficacy of SSP.

\textbf{Comparison Methods.} Six conventional fusion-based approaches i.e., AVSBench \cite{zhou2022audio}, AQFormer \cite{huang2023discovering}, Catr \cite{li2023catr}, AuTR \cite{liu2023audio}, AVSegFormer \cite{gao2024avsegformer}, Save \cite{li2025audio} and eight promt-based approaches, i.e., SAMA-AVS \cite{liu2024annotation}, MUTR \cite{yan2024referred}, GAVS \cite{wang2024prompting}, AAVS \cite{ma2024stepping}, COMBO \cite{yang2024cooperation}, TeSO \cite{wang2025can}, AVS-Mamba \cite{gong2025avs}, TPNet \cite{wang2025transformer} are chosen for comparison.

\textbf{Datasets.} The experiments make use of three datasets: two from the AVSBench-object collection, designated as v1s and v1m, along with one from the AVSBench-semantic collection, referred to as AVSS. The v1s dataset, known as the Single-source (S4) dataset, consists of 4,932 videos, which are partitioned into training (3,452), validation (740), and test (740) subsets. The v1m dataset, categorized as Multi-source (MS3), contains 424 videos, divided into training (296), validation (64), and test (64) sets. The AVSS dataset enhances the original AVSBench-object by incorporating a third subset of semantic labels that offer semantic segmentation maps as annotations \cite{zhou2024audio}. This dataset includes a total of 12,356 videos, distributed across training (8,498), validation (1,304), and test (1,554) subsets.

\textbf{Implementation Details.} The training epochs of the S4 and MS3 datasets are 30, the batch size is 2, the learning rate is set to be 1e-3 initially, and it will decay to 1e-4 at the 15th epoch; The training epochs of the AVSS dataset are 60, with the same batch size and initial learning rate, and the learning rate will decay to 1e-4 at the 20th epoch.

\textbf{Evaluation.} We employ two evaluation metrics, specifically the F-score and mean intersection over union (mIoU), to assess the models' performance. The F-score quantifies a model's accuracy by taking into account both precision and recall, thereby achieving a balance between the two. Meanwhile, mIoU evaluates the degree of overlap between the predicted segmentation and the ground truth segmentation. Higher values of the F-score and mIoU reflect robust classification abilities and effective segmentation accuracy.

\begin{table}[!t]
\centering
\caption{Model performance comparison on three datasets (best and second-best results are \textbf{bolded} and \underline{underlined}, with \% omitted).}\label{comparative}
\centering
\resizebox{0.5\textwidth}{!}{
\begin{tabular}{ccccccccc}
\midrule
\multirow{2}{*}{Method} & \multirow{2}{*}{Audio-backbone} & \multicolumn{1}{c|}{\multirow{2}{*}{Visual-backbone}} & \multicolumn{2}{c}{S4} & \multicolumn{2}{c}{MS3} & \multicolumn{2}{c}{AVSS} \\ \cmidrule{4-9} 
                        &                                 & \multicolumn{1}{c|}{}                                 & mIoU      & F-score    & mIoU      & F-score     & mIoU      & F-score      \\ \midrule
\multicolumn{9}{c}{Fusion-based Methods}                                                                                                                                                        \\ \midrule
AVSBench \cite{zhou2022audio}               & VGGish                          & \multicolumn{1}{c|}{PVT-v2}                           & 78.7      & 87.9       & 54.0      & 64.5        & 29.8      & 35.2         \\
AQFormer \cite{huang2023discovering}               & VGGish                          & \multicolumn{1}{c|}{PVT-v2}                           & 81.6      & 89.4       & 61.1      & 72.1        & -         & -            \\
Catr \cite{li2023catr}                   & VGGish                          & \multicolumn{1}{c|}{PVT-v2}                           & 81.4      & 89.6       & 59.0      & 70.0        & 32.8      & 38.5         \\
AuTR \cite{liu2023audio}                   & VGGish                          & \multicolumn{1}{c|}{Swin-Base}                        & 80.4      & 89.1       & 56.2      & 67.2        & -         & -            \\
AVSegFormer  \cite{gao2024avsegformer}           & VGGish                          & \multicolumn{1}{c|}{PVT-v2}                           & 82.1      & 89.9       & 58.4      & 69.3        & 36.7      & 42.0         \\
Save(256) \cite{li2025audio}              & VGGish                          & \multicolumn{1}{c|}{SAM}                              & 84.0      & 90.5       & 64.1      & 71.2        & -         & -            \\
Save(1024) \cite{li2025audio}             & VGGish                          & \multicolumn{1}{c|}{SAM}                              & 85.1     & 91.2       & 67.0     & 77.7        & -         & -            \\ \midrule
\multicolumn{9}{c}{Prompt-based Methods}                                                                                                                                                        \\ \midrule
SAMA-AVS \cite{liu2024annotation}               & VGGish                          & \multicolumn{1}{c|}{ViT-Huge}                         & 81.5      & 88.6       & 63.1      & 69.1        & -         & -            \\
MUTR  \cite{yan2024referred}                  & VGGish                          & \multicolumn{1}{c|}{ViT-video-Huge}                   & 81.5      & 89.8       & 65.0      & 73.0        & -         & -            \\
GAVS  \cite{wang2024prompting}                  & VGGish                          & \multicolumn{1}{c|}{ViT-Base}                         & 80.1      & 90.2       & 63.7      & 77.4        & -         & -            \\
AAVS  \cite{ma2024stepping}                  & VGGish                          & \multicolumn{1}{c|}{Swin-Base}                        & 83.2      & 91.3       & \underline{67.3}      & 77.6        & \underline{48.5}      & \underline{53.2}         \\
COMBO \cite{yang2024cooperation} & VGGish                          & \multicolumn{1}{c|}{PVT-v2}                        & 84.7      & 91.9       & 59.2      & 71.2        & 42.1      & 46.1        \\

TeSO   \cite{wang2025can}                 & VGGish                          & \multicolumn{1}{c|}{Swin-Base}                        & 83.2      & \textbf{93.3}       & 66.0      & \underline{80.1}        & 38.9      & 45.1         \\
AVS-Mamba \cite{gong2025avs}               & VGGish                          & \multicolumn{1}{c|}{PVT-v2}                           & \underline{85.0}      & \underline{92.6}       & 68.6      & 78.8        & 39.7      & 45.1         \\
TPNet  \cite{wang2025transformer}                 & VGGish                          & \multicolumn{1}{c|}{PVT-v2}                           & 82.9      & 90.8       & 59.9      & 70.9        & -         & -            \\ \midrule
SSP                     & VGGish                          & \multicolumn{1}{c|}{Swin-Base}                        & \textbf{85.4}         & \textbf{93.3}          & \textbf{72.3}         & \textbf{84.6}          &\textbf{50.1}        & \textbf{54.5}     \\ \midrule      
\end{tabular}}
\raggedright
\footnotesize \textit{note:} Save(256) and (1024) indicate the image resolution. ``-" are the results that can not be retrieved.
\end{table}
\subsection{Comparative Study}
The performance of the models on the S4, MS3, and AVSS test sets, evaluated using mIoU and F-score, is presented in Tab. \ref{comparative}. The best and second-best results are bolded and underlined. The AVS task is centered on isolating objects within audio and visual data, requiring the model to identify relevant segments. The AVSS task necessitates semantic classification for each segmented region. Consequently, both fusion-based and prompt-based methods demonstrate commendable performance on the S4 and MS3 datasets, yet they produce suboptimal results on the more challenging AVSS dataset. SSP reaches the state-of-the-art method. It consistently outperforms the baseline models, AAVS, achieving improvements of 2.2\% and 1.9\% on S4, 5.0\% and 7.0\% on MS3, and 1.6\% and 1.3\% on the AVSS dataset, correspondingly on mIoU and F-score.

Compared to fusion-based methods, prompt-based approaches are effective because they provide supplementary contextual information. These prompts enhance the model's scene understanding by offering specific details that guide the segmentation process, allowing for better differentiation between objects and their attributes, and ultimately improving performance in tasks requiring nuanced interpretation of audio-visual data. However, when utilizing prompts as supplementary information, it is crucial to assess the effectiveness of alignment between modalities.
To avoid misalignment, we use a more direct prompt in SSP, the optical flow, intervening in the visual data, which contains the segmented objects through the pre-mask technique. While optical flow effectively detects moving objects, it may exclude sound-emitting stationary objects or include noise from moving but silent objects. This limitation can impact the model's performance by either omitting relevant segmented objects or introducing unwanted noise. We mitigate this by integrating textual prompts and aligning them with visual features by VTA to enhance the original visual representation. Having the dual assist prompts, SSP is a multi-faceted approach that can ensure a more robust segmentation process, leveraging both dynamic and contextual insights.

\subsection{Ablation Study}

\begin{table}[!t]
\centering
\caption{Ablation Studies on three datasets. We evaluate the impact of adding key modules from our proposed framework.}\label{ablation}
\resizebox{0.48\textwidth}{!}{
\begin{tabular}{c|cccccc}
\midrule
\multirow{2}{*}{Component} & \multicolumn{2}{c}{S4} & \multicolumn{2}{c}{MS3} & \multicolumn{2}{c}{AVSS} \\ \cmidrule{2-7} 
                           & mIoU     & F-socre     & mIoU      & F-socre     & mIoU      & F-socre      \\ \midrule
AAVS  \cite{ma2024stepping}                    &   83.2       &   91.3          & 67.3          &   77.6          &   48.5        & 53.2  \\
\textit{w.} Pre-mask                 & 84.1         &   92.3          &   69.5        &    82.3         &   49.2        &  53.7            \\
\textit{w.} Pre-mask \textit{w.} Post-mask                 & 85.0         &   92.5          &   70.2        &    83.7         &   49.4        &  53.9            \\
\textit{w.} Textual prompts \textit{w/o.} VTA  &       83.7   &      91.8       &  68.1         & 78.9             &  48.6         &  53.0            \\
\textit{w.} Textual prompts \textit{w.} VTA    &        84.3    &   92.0        &   69.8        & 83.3             &  49.0         &   53.4           \\
\textit{w/o.} Post-mask              &   84.5       &  92.4           &  70.4         &  83.2           &   49.7        &   54.1           \\ \midrule
Full model                 &  \textbf{85.4}        &  \textbf{93.3}           &   \textbf{72.3}        &   \textbf{84.6}          & \textbf{50.1}          &   \textbf{54.5}           \\ \midrule
\end{tabular}}

\raggedright
\footnotesize \textit{note:} VTA must be used in the presence of the textual prompts. 
\end{table}

In this section, we present ablation studies in Tab. \ref{ablation} to evaluate the impact of four key components within SSP: (1) pre-mask technique with the optical flow, (2) two textual prompts, (3) VTA module, and (4) post-mask technique with the optical flow.

\begin{figure}[!t]
  \centering
    \begin{minipage}{0.23\textwidth}
    \centering
    \includegraphics[scale=0.3]{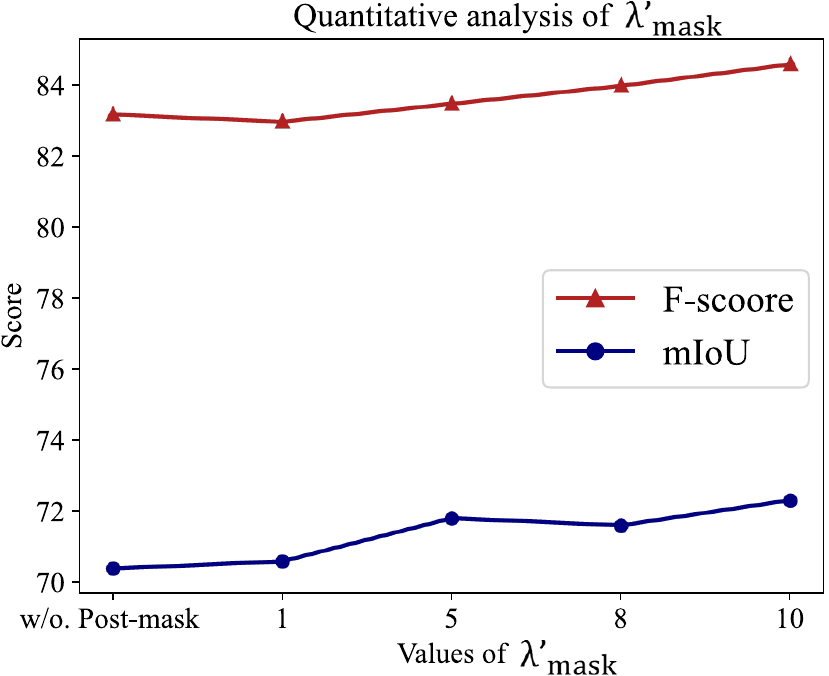}
    \caption{Five values of $\lambda'_\text{mask}$ assigned to $\mathcal{L}'_\text{mask}$, evaluating the impact to the overall performance using the MS3 dataset.}
    \label{lambdaQuali2}
  \end{minipage}
  \hfill
  \begin{minipage}{0.235\textwidth}
    \centering
    \includegraphics[scale=0.30]{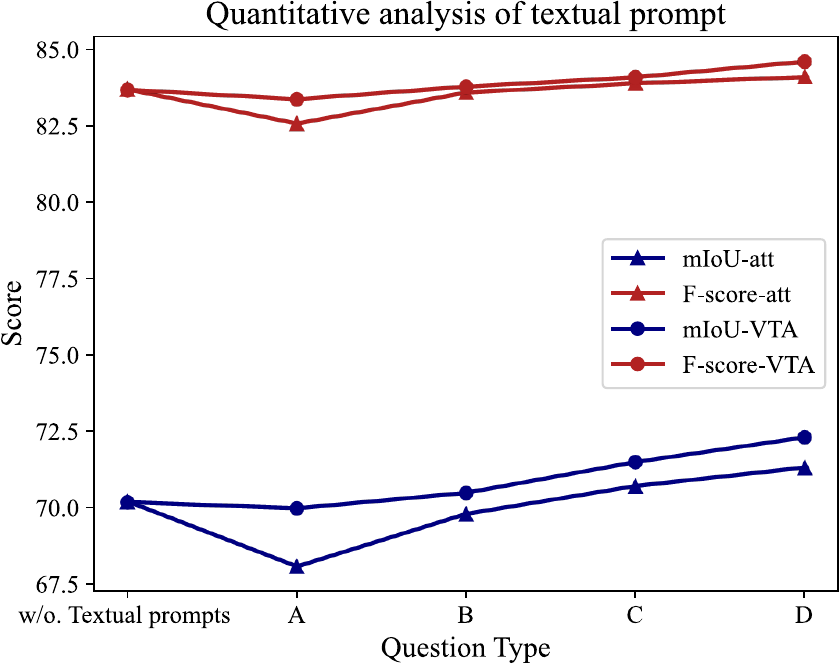}
    \caption{Evaluating the quality of the textual prompts by ranking on the mIoU and F-score using the MS3 dataset.} 
    \label{promptQuali1}
  \end{minipage}
\end{figure}

Two key insights in Tab. \ref{ablation} are worth noting:

1. When solely relying on optical flow as a prompt, the integration of the pre-mask technique yields significant performance. Specifically, the mIoU and F-score for the S4, MS3, and AVSS datasets have improved by 0.9\%, 1.0\%; 2.2\%, 4.7\%; and 0.7\% 0.5\%, respectively. This improvement can be attributed to optical flow's ability to effectively identify the majority of moving and sound-producing objects, allowing the model to apply the mask to the original image prior to encoding. Consequently, this enables the model to step upon a more refined foundation, facilitating further mask refinement. With the integration of both pre-mask and post-mask techniques, the model's accuracy approaches SOTA levels. This further substantiates the effectiveness of optical flow as a prompt in assisting the AVS task. Different values of $\lambda'_\text{mask}$ assigned to $\mathcal{L}'_\text{mask}$ rank on the MS3 dataset in Fig. \ref{lambdaQuali2} (Full model settings) further highlights the necessity for the model to be compelled to learn this feature prior to the effective utilization of optical flow when the $\mathcal{M}_\text{GT}$ is excluded during testing. Establishing a solid foundational understanding can leverage the dynamic insights provided by optical flow to enhance performance.

2. When using only textual prompts without the VTA integration, we implement a simple cross-attention mechanism \cite{huang2019ccnet}, denoted as ``att" in Fig. \ref{promptQuali1} (settings in \textit{w/o.} textual prompts are the utilization of pre-mask and post-mask techniques. In A-D, we present the full model utilizing different integration modules: ``att" and VTA. The results show that VTA achieves an average improvement of 1.1\% in mIoU and 0.5\% in F-score compared to the 'att' module, demonstrating the effectiveness of integrating visual and textual information through VTA. Furthermore, variations in the quality of the textual prompts significantly impact model performance, with the best (D) and the worst (A) configurations yielding differences of 2.3\% and 1.2\% in mIoU and F-score with VTA, and 3.2\% and 1.5\% using the ``att" module. In the case of configuration A, where the prompt quality is low, a substantial amount of redundant information is introduced. When combined with the simplistic attention structure, this redundancy exacerbates the degradation in performance. To further explore the impact of quality on performance, we conduct a quantitative analysis.

\begin{figure*}[!t]
  \centering
  \makebox[1\textwidth]{\includegraphics[scale=0.50]{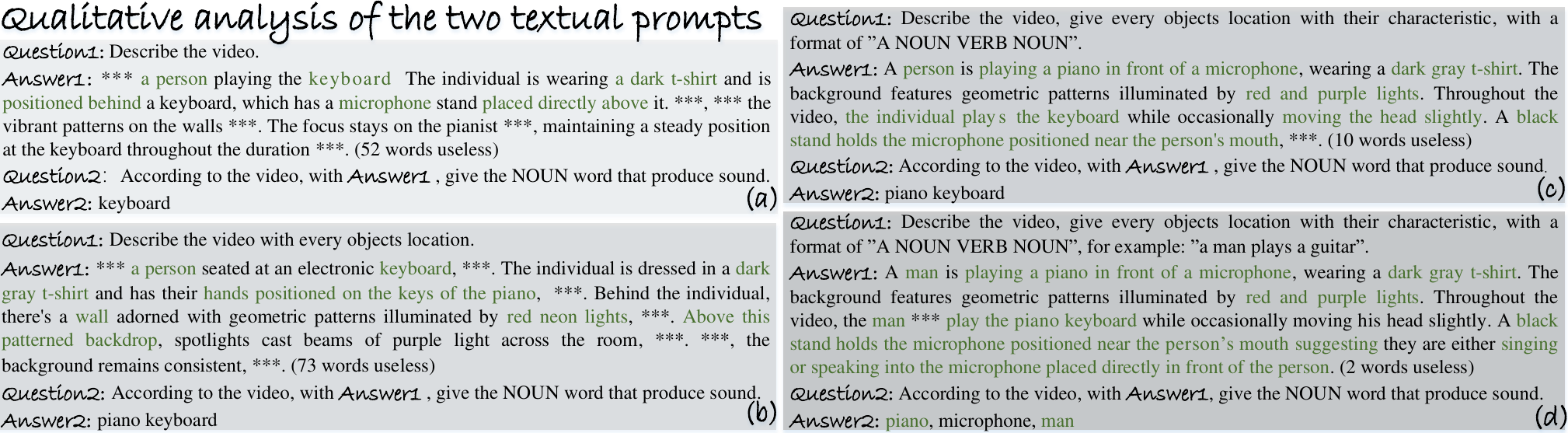}}
  \caption{This qualitative analysis of four different questions (a, b, c, and d) affects the generation quality of the answers (textual prompt) quality, the results demonstrate a gradually refining trend. $\text{Q}_1^{\text{a-d}}$ has a progressive refinement trend, with the corresponding answer $\text{A}_1^{\text{a-d}}$ containing more useful (text in green) and less useless (``***") information. }\label{promptQuan}
\end{figure*}
\subsection{Qualitative Analysis}
\begin{figure*}[!ht]
  \centering
  \makebox[1\textwidth]{\includegraphics[scale=0.195]{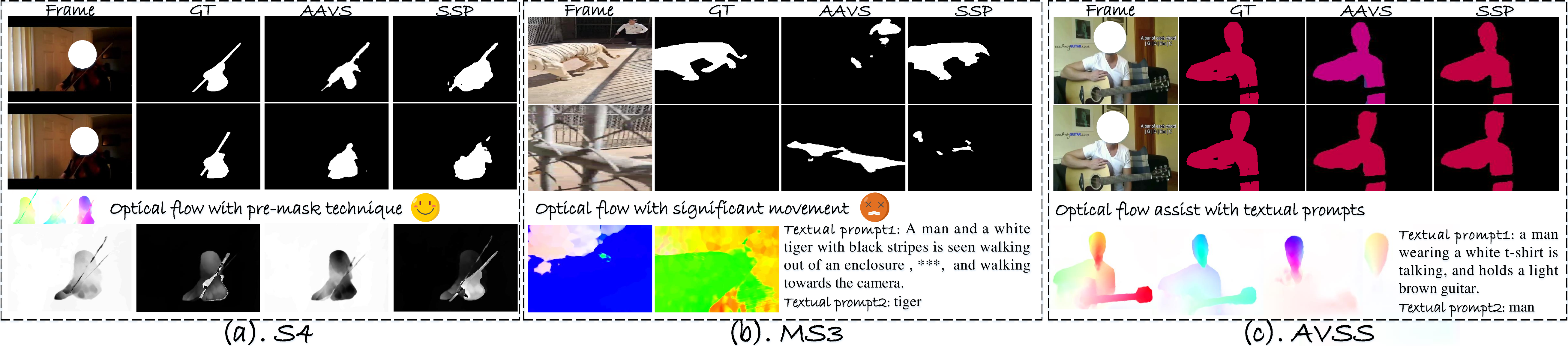}}
  \caption{Visualization results using the S4, MS3, and AVSS datasets in (a), (b), and (c) respectively, we aim to illustrate the assistance role of the optical flow and the textual prompts. By comparing configurations (a) and (b), the optical flow effectively captures the contours of the sound-producing, violin. However, in (b), the significant movement towards the tiger introduces redundant information, hindering the optical flow's performance. In this context, the two textual prompts play compensatory roles. In (c), the optical flow can outline both the man and the guitar. However, the man is speaking with the silent guitar, the textual prompts again serve to compensate for this discrepancy.}\label{vis}
\end{figure*}

To understand how quality affects performance, we conduct the experiment using the MS3 dataset, results are presented in Fig. \ref{promptQuan}, which features four types of $\text{Q}_1$ (a, b, c, and d) that follow a progressive trend. Each successive prompt, $\text{A}_1$, contains progressively less redundant information and provides a more precise description of the video. Since $\text{A}_1$ serves as the foundation for $\text{A}_2$, excessive irrelevant information in $\text{A}_1$ results in similarly unhelpful outputs in $\text{A}_2$. Furthermore, when comparing $\text{Q}_1^\text{c}$ and $\text{Q}_1^\text{d}$, the answer $\text{A}_1^\text{d}$ includes specific descriptions of the \textless\textit{individual}\textgreater and the \textless\textit{person}\textgreater in $\text{A}_1^\text{c}$, identifying them as \textless\textit{man}\textgreater, due to the absence of specific contextual information in $\text{Q}_1^\text{c}$, the MLLM typically resorts to more neutral terms when handling vague descriptions \cite{nie2024mmrel, kawabe2024duetml}. Having a more precise $\text{A}_1^\text{d}$, it can successfully allow for the identification of another sound-producing object, \textless\textit{man}\textgreater. 

\subsection{Visualization Results}
In our latest analysis, we present the visualization results in Fig. \ref{vis}, which provide a visual comparison between SSP and AAVS using the S4, MS3, and AVSS datasets in (a), (b), and (c), respectively. We aim to highlight the roles of assistance between the optical flow and the two textual prompts, particularly in the last column.

By comparing (a) and (b) in the final column, we observe that the optical flow in (a) effectively captures the contours of the sound-producing object, aided by the integration of the GT mask within the pre-mask technique. In contrast, in (b), the object generating sound is a tiger, which exhibits significant movement. This rapid motion prevents the optical flow from capturing valuable information for subsequent visual data analysis. Consequently, the textual prompts play a compensatory role, providing clarity by identifying the sound-producing object as \textless\textit{tiger}\textgreater.  Turning to (c), the optical flow adeptly identifies a man holding a guitar. However, since the man is conversing with the guitar remains silent, the textual prompt plays a crucial role in filtering out the guitar. This results in a precise segmentation of the man, isolating him from the surrounding elements.

\section{Conclusion}
\label{sec:conclusion}
This paper has presented SSP, an architecture that uses additional prompts to enhance AVS and AVSS tasks. By integrating optical flow in the pre-mask technique, SSP improves segmentation accuracy with compensatory textual prompts for stationary, sound-producing objects. The VTA module offers aligned features for visual embeddings, while the post-mask technique boosts learning effectiveness. Empirical results demonstrate SSP's superiority, establishing it as a SOTA method. This work sets the stage for more robust results, with future research focused on using more nuanced prompts for further improvement.

{
    \small
    \bibliographystyle{ieeenat_fullname}
    \bibliography{main}
}

\end{document}